# Exploration of Deep Learning Based Recognition for Urdu Text

Sumaiya Fazal
*Dept. Computer Science*
*FAST-National University* Pakistan
p200401@nu.edu.pk

Sheeraz Ahmed
*Dept. Computer Science*
*IQRA National University* Pakistan
Sheeraz.ahmad@inu.edu.pk

*Abstract*—**Urdu is a cursive script language and has similarities with Arabic and many other South Asian languages. Urdu is difficult to classify due to its complex geometrical and morphological structure. Character classification can be processed further if segmentation technique is efficient, but due to context sensitivity in Urdu, segmentation-based recognition often results with high error rate. Our proposed approach for Urdu optical character recognition system is a component-based classification relying on automatic feature learning technique called convolutional neural network. CNN is trained and tested on Urdu text dataset, which is generated through permutation process of three characters and further proceeds to discarding unnecessary images by applying connected component technique in order to obtain ligature only. Hierarchical neural network is implemented with two levels to deal with three degrees of character permutations and component classification Our model successfully achieved 0.99% for component classification.**

*Index Terms*—**Deep learning, Urdu OCR, Convolutional Neural Network, Connected Component**

## I. Introduction

Urdu is national language of Pakistan and recently declared as official language. It is a popular and widely spoken language of Pakistan, India, Bangladesh and other South Asian countries along UK and Canada. Almost, 11 million speakers of Urdu are in Pakistan and 300 million and above are in the entire world [1]. Urdu follows two different writing styles Naskh and Nastaliq. Urdu writing is based on Nastaliq style however, Arabic is based on Naskh style of writing which is different from Nastaliq in terms of diagonality [2]. Urdu is rather complex as its morphology and syntax structure foundations are mixture of Persian, Sanskrit and Arabic [3].

Field of pattern recognition is transformed in recent years and newly developed systems produced some phenomenal results in the filed of image recognition. Western languages are ahead of deploying updated systems as they are rich-resource languages due to the wide range of work, while same is not true when it comes to South Asian languages, resource scarcity is one of the major setback behind a slow research in Urdu OCR as not enough data is available for training purpose, and Urdu language also have more limitations and complexities of defining a word or character as compared to any other language which imposes a barrier to find a suitable technique. Challenges and complexities in implementation of

Urdu Nastaliq recognition [4] are briefly explained in various papers.

Digitization of printed Urdu text takes input text and tokenizes it into words and characters, in case of English character, tokenization is an easy process because spaces at the end of word defines a boundary. There are many Asian languages which are probably similar such as Thai, Khmer, Lao and Dzongkha but does not offer word boundaries and therefore avoids the use of white spaces to mark word endings [5]. Segmentation in Urdu language is a complex procedure it faces space omission and space insertion error.

Character and ligature recognition of printed Urdu documents passes through segmentation procedure first and segmentation of Urdu is not only a complex procedure, the segmentation based recognition often results with high error rate, recognizing a whole word in Urdu OCR is yet unaccomplished, due to this reason mostly researches are moving to the segmentation free methods where recognition is apparently easy, character, ligature and sentence based recognitions are providing good results and with diversity in innovative research the researchers have shift their focus on ligature and sentence based OCR systems. Ligature recognition through feature learning produced some promising results but manually hand-engineered features requires domain expertise and need to be carefully designed. A prevalent feature extractor technique for better learning needs to be used to obtain high results.

Our proposed approach is component-based method trained on Urdu text which identifies each ligature as a separate component and recognizes it whether component is combination of one character, two or three. Component separation is done by using two pass connected component algorithm [6] to reduce the number of classes, which label pixels based on their neighbor pixels, if pixels belong to the same region same label is assigned, after separating un-necessary components, convolutional neural network is used for learning and classification. Learning and classification is further divided into two levels where at level-0 we classify the number of character in a component and at level-1 which holds degree of characters individual components are being recognized. Convolutional neural network is deep learning architecture which have property of automatic feature learning, instead of defining features manually it automatically extracts features based on weights. Dataset of Urdu ligature is a text-based ligature generated through permutation for three characters and trained through statistical technique of deep learning. Result we achieved at level-0 is 0.99%.

## II. Background

### A. Urdu Optical Character Recognition

OCR (Optical Character Recognition) process is based on scanning document and converting it into a text file to make it editable but OCR process will produce more efficient results if word or characters are defined by some boundary. Character segmentation is a critical area in OCR due to the fact that isolated characters have high recognition as compare to character in word. Space at the end of a word defines a boundary but not all languages follow similar rule or agreement for defining a word boundary and Urdu is one of those languages. Segmentation strategies are divided into three groups first is classical approach which is mainly based on dissecting or cutting image into understandable components if it fulfills all the required properties of character, second segmentation strategy is recognition based and third is holistic method.

Segmentation of Urdu words is a complex task, and segmentation based recognition is mostly prone to high error for variety of classifiers, while segmentation free recognition [7] for Arabic text produced good results through multilayer perceptron network with back-propagation learning, where at feature extraction stage a set of moment invarients descriptors which are invarient under shift, scaling and rotation are used.

Mandana et al. [8] connected component recognition is used where rectangular is created to group the connected regions. Skew detection and correction is used to determine angle of each group and for feature extraction contour tracing is further explained. U.pal et al. [9] character recognition is performed through binary tree classifier and for feature extraction topological, contour and water reservoir methods are combined. Accuracy achieved is 97.8% for characters only.

Nabeel et al. [10] approach for isolated characters recognition first analyzes the geometric properties as well as visual and hand-crafted set of features. Set of features consists of primary stroke features and secondary stroke features and then weighted linear classifier is used to perform classification.

Malik et al. [11] a holistic approach for classification is used, feature extraction of word is based on structural and gradient features and SVM is used for classification. SVM performs better as compare to other classifiers but in segmentation based method recognition is difficult for SVM. Sajjad et al. [12] achieved high accuracy through convolutional neural networks on hand written Arabic and Farsi digits. Automatic extraction of Farsi digit features performed by CNN and rejection strategies were used to find out the samples which are hard to recognize.

Ali et al. [13] this approach is divided into indexing and retrieval, in indexing from printed image PWs are extracted these PWs represents features, feature extraction is based on hand-engineered features, in next phase query word is compared with indexed PWs for word retrieval. Khalil et al. [14] two databases train and test are used, eigen values and eigen vectors are calculated for images, a feature vector and a threshold value is selected to define max distance between images of both datasets, comparison is based on the identified characters are matched with train dataset. Nazlay et al. [15] developed an application Nabocr is an OCR system specifically used for Arabic script, initially it trains raw Arabic script and outputs a dataset which consists of ligature in form of feature vector, it takes an image as input and convert image into text by using dataset of ligatures, third part of application is user interface it enables users to edit recognition output.

Shazia et al. [16] sentence boundary identification technique uses feed forward neural network for the identification of sentence terminators in Urdu text files, corpus converted into bipolar descriptor array by using word-tag frequency, arrays are used as training set for FFNN. It can be trained on small datasets, no hand-crafted rules needed, and it requires less storage, precision achieved up to 93.05 % and re-call 99.53 %. Safia et al. [17] connected components were extracted and divided into primary and secondary ligatures, each ligature represents a set of features where all the ligatures are clustered in groups, these clusters are used as training data and HMM is used for each ligature. After primary ligature recognition, secondary is recognized for completion of word. Tofik et al. [18] OCR system performed on 23204 Urdu ligatures, after feature extraction the corresponding classes obtained were 1600. ID is associated with each feature point through calculating descriptors, features are extracted based on hand-crafted rules. Ligature in digital input image is converted to editable text by matching the ligature with highest descriptor ID.

Ibrar et al. [19] a deep learning network stacked denoising autoencoder for automatic feature extraction of ligature image pixel values used for Urdu OCR. First deep stack of denoising

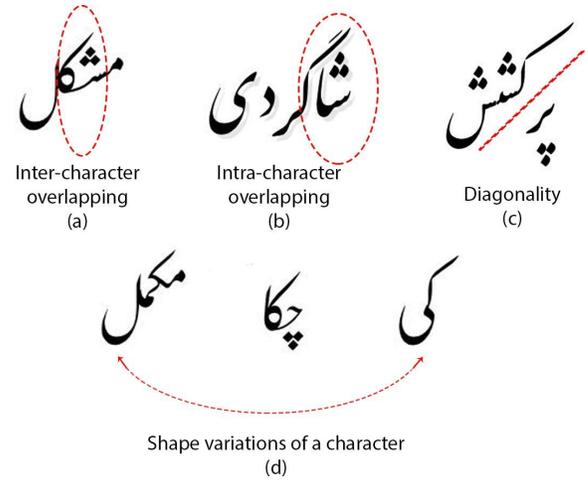

Inter-character overlapping
(a)

Intra-character overlapping
(b)

Diagonality
(c)

Shape variations of a character
(d)

Fig. 1. Complexities in Urdu language.

autoencoders are pre-trained on Urdu ligature images in an unsupervised manner layer by layer then this trained layer is forwarded to the MLP which fine-tunes it through back propagation and classifies it through logistic regression. Accuracies reported are 93% to 96%.

Naila et al. [15] system has dataset of 2430 ligatures, initially images are fed to the system, segmentation is performed by horizontal projection for line segmentation, for ligature segmentation vertical projection is used. Feature extraction is performed by hand crafted rules. Once features are extracted agglomerative hierarchical clustering is used. Total of 417 clusters were generated, 4 classification techniques are used. Accuracy of decision tree, linear discrimination, naive bayes, KNN were 62%, 61%, 73%, 100% respectively.

Ibrar et al. [7] implemented a gated bidirectional long short-term memory which is based on recurrent neural network-LSTM for recognition purpose. The method extracts features from raw pixel, first model is trained on noise free version of images and tested by feeding degraded images, accuracy achieved is 96.71%. Saeeda et al. [20] hybrid approach proposed using CNN and MDLSTM, first CNN extract lower level features from MNIST database to train the network, and applies six filter along contour filter which produces contour image and filtered images which works as an input for MDLSTM and uses random weights which maps feature into a lower dimensional space, concatenation of all vectors is performed for CTC layer.

Saeeda et al. [21] MDLSTM technique, sliding window approach is used to divide text line into characters and divide each in frame, extracts features from frames and creates feature vector sequence. In second stage, multi-dimensional long short-term memory is used. Final stage CTC is used to recognize character by using conditional probability of label with high value. The proposed approach significantly outperforms the state-of-the-art Urdu recognition system with results of training error 3.72 % and recognition accuracy is 94.97 %.

### B. Convolutional Neural Network

Data can vary basis on internal patterns and structures, to learn such data representation, model is used to create multiple levels of abstraction by forming different levels of layers and extracting data representation at each layer. Deep learning made some phenomenal improvements in the field of speech-recognition, object detection and image processing such as OCR of hand-written documents, also numerous other domains such as bioinformatics, robotics, identifying tumors in medical diagnosis, drug discovery and security purpose. Pattern-recognition requires feature engineering which can be detected by a classifier due the need of conventional machine learning methods which processes data in raw form, careful feature-engineering expertise must be considered first.

Deep learning employs data representation methods, initially an image which is an array of pixels or raw input is fed to the classifier, at first layer low level representation is extracted such as edges, next layer extracts motifs another layer detect parts of object and at final layer object is recognized. The main feature behind representation learning is automatic feature detection, no human expertise is required. Deep learning automatically learns features and recognizes object by adjusting the internal parameters called weights and reduces the error. Weight vectors are adjusted through gradient vector, miner change in a weight increases or decreases the error. In practical many optimization techniques are used to compute errors and adjusts the weights. Repeating the process on training set, test is used to measures the performance of system, which produces result based on the ability of system for the inputs unknown during the training.

ConvNet works on four key ideas [22]: layers, pooling, weights and local connections. The structure of ConvNet is composed of stages. Initial stages consist of convolutional layers and maxpool layers along weights called filter, different number of filters passes over the single image and creates a feature map, sum of the weights passes through the activation function such as ReLU, or LeakyReLU to squashing the input value into a range. Convolutional layer detects conjucted features of previous layer and semantically similar features are merged by pooling layer, max pool layer fetches a higher or maximum bit of a patch and reduces the dimensions. Backpropagation end-to-end supervised learning algorithm is used to train all the weights of every filter in convolutional neural network. Convolutional neural network takes an image as a tensor, image dimension is its width and height which represents number of pixels and depth referred as channels. A patch of an image is passes through the filter smaller than the image size, the dot product of both filter and a patch outputs a value which represents a match of a pattern. Strides are the steps which filter takes while traversing the image, larger stride will compute faster and generate small activation map as compared to smaller strides.

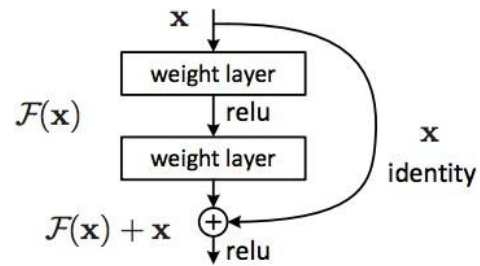

Fig. 2. Resiudal Nets.

Deep residual learning framework [23] is simpler formation of deep neural networks, layers are being reformulated with reference of input layers. It handles the degrading problem caused by deep layer network where eventually accuracy degrades after getting saturated. Residuals solves the degradation problem, faster training then stacked deep network and higher accuracy due to lower number of parameters. Hundred layers of deep network represents complex function it helps in understanding many features at abstract level, however, it does not always helpful with causing vanishing gradient problem. Resnets (See Fig.2) adds shortcut to mainstream stacked network and allows one of the block to learn an identity function and skip connection helps with vanishing gradients.

## III. METHODOLOGY

### A. Component Based Identification

As per previous approaches printed Urdu text was based on character classification and ligature classification but development in pattern-recognition technology transform the main-stream hand-crafted feature ideology to the automatic feature learning. Our system is entirely based on component-based identification. State-of-the art segmentation methods often lead to the lowest recognition rate since Urdu language has problems of complex morphological structure of defining a word.

Feature learning needs a very large dataset to be trained initially, dataset creation is mainly depends on various phases. The major setback in Urdu research is resource scarcity for training purpose. The 83000 ligature dataset is the largest available dataset of Urdu OCR [24]. In our proposed method (See Fig.4) intially Naskh category fonts were selected and permutations (See Formula.1) which arranges all the characters in some order used for creation of component, in our case upto three character permutations were generated. The motivation behind using three character permutation is that the longest word in Urdu consists of probably seven characters while most of the Urdu words are combination of two or three characters only.

$$^nP_k = \frac{n!}{(n-k)!} \tag{1}$$

As a result character image of 100*100 is created, first permutation, generates images for 38 characters, second permutation creates 306 images, combination of two characters, and third permutation creates 4, 896 images through joining three characters. Number of characters can be increased depending on the need, our dataset consists of three character permutations but to train deep neural netwrok on large dataset (See Table.I) we included 15 more fonts which makes dataset of approximately 20, 000 Urdu text images.

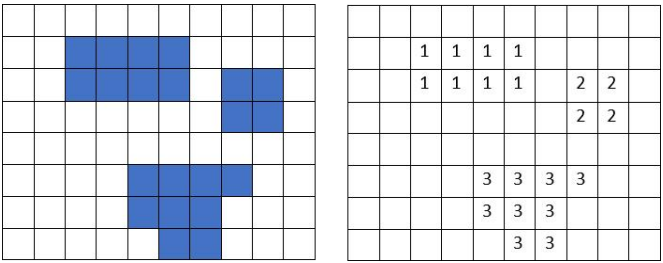

Fig. 3. Example of a figure caption.

| Dataset | No. of Classes | No. of Fonts | No. of Images |
|---|---|---|---|
| Level-0 | | | |
| Level-0 | 3 | 15 | 19, 435 |
| Level-1 | | | |
| Degree-1 | 19 | 15 | 285 |
| Degree-2 | 150 | 15 | 2, 250 |
| Degree-3 | 1260 | 15 | 18, 900 |

Connected component is a methodology where a set of pixels which relates to each other by means of neighborhood assigned with same class (See Fig.3), a connected component algorithm, labels all the connected pixels in an image with a unique label if it belongs to the same component. Various connected-component techniques are part of pattern matching and recognition and have produced some tremendous results such as stroke-width transformation algorithm [25] and two pass connected component algorithm [6] are improved versions of connected-component labeling algorithm.

Our proposed methods uses connected-component which separates all those characters, dot and diacritics which are not joined, it separates dots and diacritics from words and discards them, after removing dots and diacritics from all the permutations, characters which were not joined were discarded as well, such as, at second and third permutations many characters can not join due to the context sensitivity and will be discarded, for 38 characters dataset, after removing dots and diacritics common characters were removed as well and dataset is left with 19 characters only.

Finally dataset is consists of join characters or components only, after discarding un-necessary images our dataset is reduced to 20, 000 images for three character permutations, to increase the dataset image augmentation process is also used in architecture for better learning purpose.

### B. Architecture

ConvNets are form of neural networks that have learnable weights with multiple hidden layers and properties encoded in architecture which reduces number of parameters of network. ConvNets are built upon sequence of layers, Convolutional Layer, Pooling layer and Fully-connected layer.

Input layer: It represents our image dimensions and channels, width and height tells us about number of pixels an image can have.

Conv2D: It acquires pixel value of images as a matrics such as [100,100,1] width 100, height 100, and channel of image which in our case is 1. It also consists of number of filters and filter size which computes a dot product of input under region of filter size. Many layers of Conv2D with various sizes and filters are part of our model.

Activation layer: RELU, LeakyReLU and SRELU are activation functions to perform thresholding while SoftMax is used at the end layer to produce results with higher probability.

Maxpool layer: Pool layer down samples and takes max of each feature map which results in form of higher bit informa-tion only and discards rest of the values, such as [100,100,32] will be [50,50,32], resultant activation map comprises higher information referred as features with strong correlation. It benefits us by decreasing the time processing needs and storage it occupies. Different pooling layers GlobaAveragePooling, AveragePooling available in keras can be used.

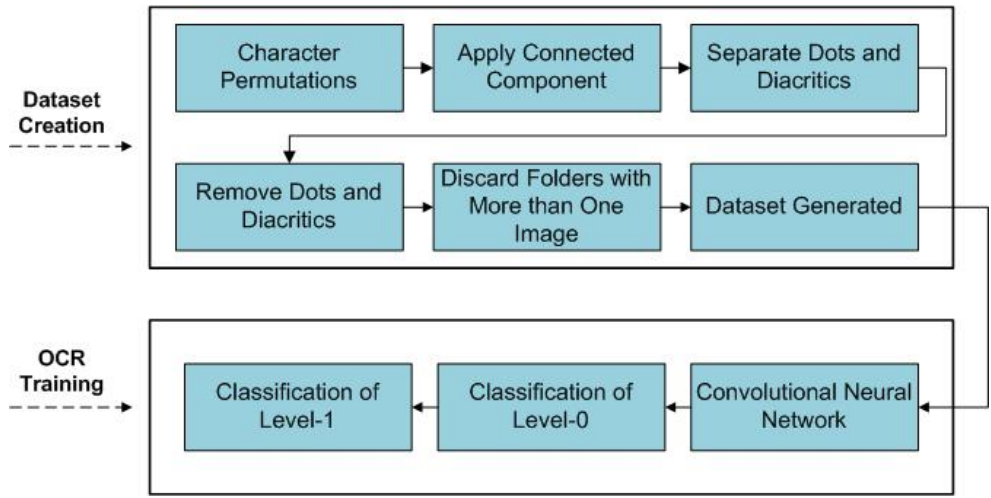

Fig. 4.  Proposed Model.

Dense layer: It is fully-connected layer results in class score such as among the 38 classes of our dataset will be separately connected to each node.

Dropout layer: Dropout is a regularization approach which adds penalty to prevent from learning interdependent weight.

Optimizers: We used RMSprop for updating weights with learning rate=0.0001.

Class weights: Level-0 contains three number of classes and number of image for all the classes are not similar which makes our training data imbalance. To perform training for imbalance dataset class weight parameter available in keras is used. Class one assigned with highest weight as compared to other two classes (See Table. IV).

Data Augmentation: ImageDataGenerator is capable of producing augmented images for better learning, it will create augmented images with various properties such as zoom, rotation at different directions and flip in no time, although memory overhead reduces but it increases training time.

### C. Hierarchical neural network

Hierarchical neural network [26] is multilevel neural network forms a tree structured architecture and can be used to perform hierarchical based classification for pattern recognition such as, character classification or ligature classification. Our proposed model implements a hierarchical neural architecture (See Fig.**??**) which trains input pixels and outputs a component based on the number of characters, we separately trained four neural networks at each level and joined them later. At level-0 datasets of all three characters of three classes are trained which consequently identifies the class, component belongs to, i.e. degree one represents component comprises one character, degree two component is combination of two characters and same demonstrates for degree three. The key focus of our paper is to classify the component based on characters. Following the hierarchy level-1 intends to show the component predicted by previous level. The output of level-0 is input of level-1, if a component identified as a class three at level-0 it will be push to the degree-3 where further classification will take place. Results achieved for all the datasets indicates effectiveness of our proposed scheme.

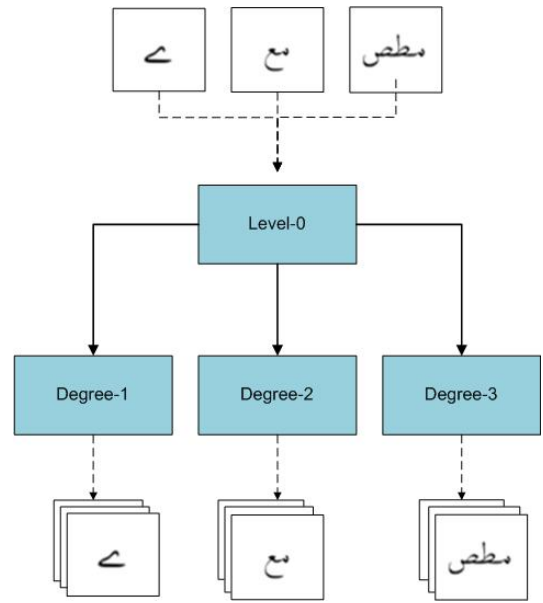

Fig. 5.  Hierarchical neural networks.

### D. Result Analysis

Deep convolutional neural network is implemented for the component recognition, keras is a high-level API which is user friendly, fast and can be used on CPU and GPU both, recognition rates (See Table.III) are given for all the levels and degree of characters, at level-0, our goal is to classify component based on character it comprises. (See Fig:6) is loss which illustrates error rates of training and validation datasets, error rate is decreasing with increasing number of epochs, number of epochs are 20, training error rate is 0.0013% while validation error rate is 0.0003%. Accuracy achieved (See Fig.**??**) for training dataset is 0.9998% and for validation its 0.99% as dataset is imbalance in this scenario for which keras, class-weight library is used to assign weights to the classes and besides accuracy, precision, recall and F-1 score is reported as well which is 0.99% in this case.

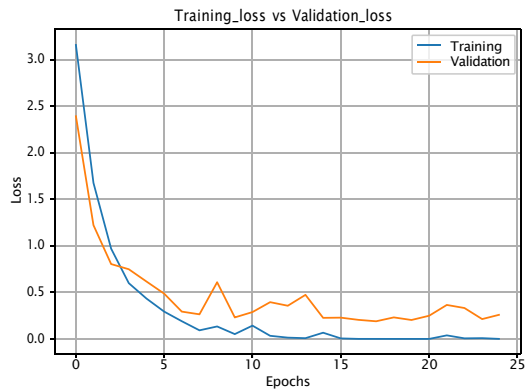

Fig. 8. Loss of Degree-1 at Level-1.

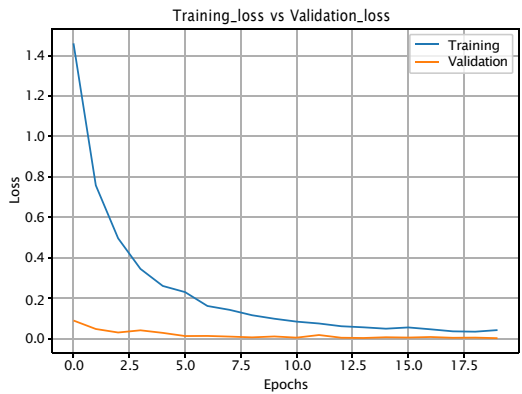

Fig. 6. Loss at Level-0

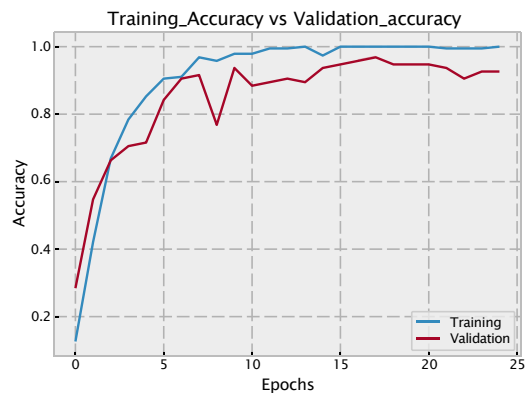

Fig. 9. Accuracy of Degree-1 at Level-1.

dropout up-to 0.25 variance in dataset is decreased, achieved validation accuracy (See Fig.11) of model is 0.98%.

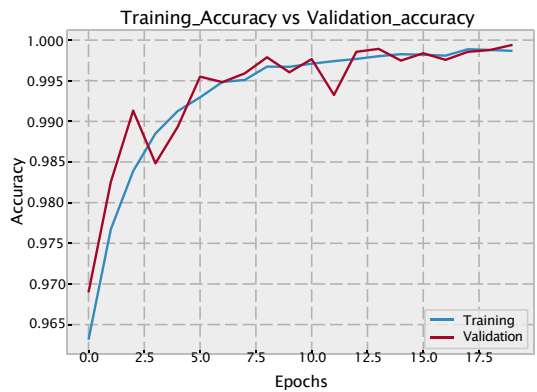

Fig. 7. Accuracy at Level-0.

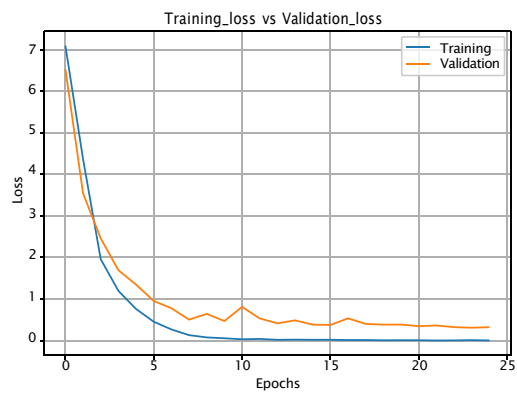

Fig. 10. Loss of Degree-2 at Level1

Four layered CNN model is employed for level-1 at degree-1 but size of filter is (8, 7, 6, 3), number of epochs are 25 and error rate is dropping at each epoch (See Fig.8), 0.0002% and 0.18% are training and validation error rates respectively, accuracy (See Fig.9) for training is 1% and for validation set is 0.97%.

At Degree of character two dataset is increased with 4, 590 images (See Table:I) of 100*100, for training set error rate is 0.005% and for validation it is 0.3% (See Fig.10), by adding a

At Degree of character three dataset is increased with 18, 900 images (See Table:I) of 100*100, with number of classes 1, 260, resnet is implemented for training set error rate is

TABLE II
CNN ARCHITECTURE CONFIGURATION

| Dataset | Epoch | Activation Layer | Learning rate | Optimizer | No. of Filters | Regularizer | Loss | Class Weights |
|---|---|---|---|---|---|---|---|---|
| Level-0 | 5 | SReLU | lr=le-3 | Adam | 48 | 0.2 | Categorical$_c$rossentropy | 1:350, 2:30, 3:10 |
| Level-1 | | | | | | | | |
| Degree-1 | 25 | SReLU | lr=le-3 | RMSProp | 128 | 0.2 | Categorical$_c$rossentropy | Nill |
| Degree-2 | 25 | SReLU | lr=le-3 | RMSProp | 128 | 0.2 | Categorical$_c$rossentropy | Nill |
| Degree-3 | 25 | SReLU | lr=le-3 | RMSProp | 128 | 0.2 | Categorical$_c$rossentropy | Nill |

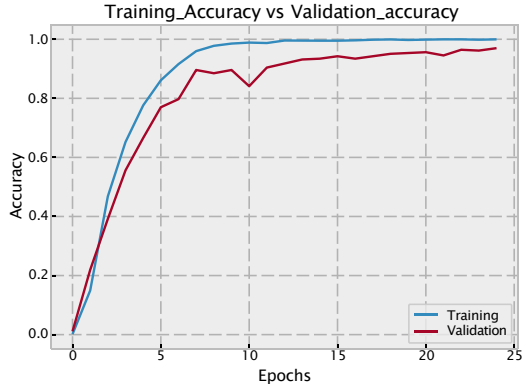

Fig. 11. Accuracy of Degree-2 at Level-1

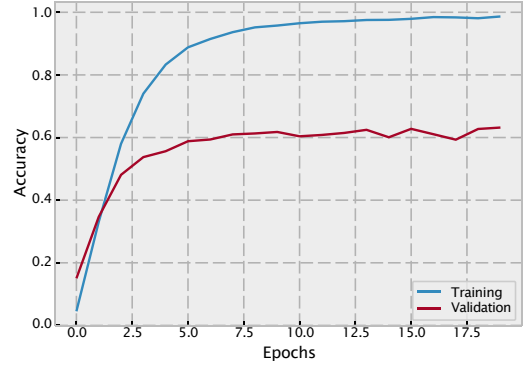

Fig. 13. Accuracy of Degree-3 at Level-1

0.003% and for validation it is 0.9% (See Fig.12), achieved validation accuracy (See Fig.13) of model is 0.63%.

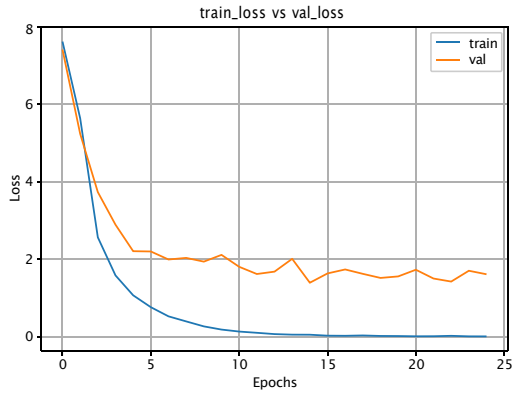

Fig. 12. Loss of Degree-3 at Level-1

TABLE III
RECOGNITION RATE

| Dataset | Accuracy | Precision | Recall | F-1 Score |
|---|---|---|---|---|
| Level-0 | 0.99% | 0.99% | 0.99% | 0.99% |
| Level-1 | | | | |
| Degree-1 | 0.968% | 0.98% | 0.97% | 0.97% |
| Degree-2 | 0.98 % | 0.94 % | 0.93 % | 0.93% |
| Degree-3 | 0.63 % | 0.72% | 0.63% | 0.63% |

## CONCLUSION

In this paper, component-based recognition for Urdu Naskh text is implemented through convolutional neural networks a deep learning-based architecture. Urdu character recognition has been a complex task to handle as per various limitations of a language, segmentation free recognition so far produced good results as compared to segmentation-based strategies.

Our proposed recognition model helps us to identify a component, based on number of characters it consists, through convolutional learning and classification which further forms a hierarchical neural network leading towards the specific component identified by the previous level, we also used residual convolutional neural networks for better learning at level-1 where dataset was small. Accuracy achieved for level-0, identification of number of components is 0.99% with F1 score 0.99%. This research is part of a huge project where targeted documents are specifically hand written Naskh font of Urdu language, while most of the existing Urdu documents are in Nastaleeq font. After this part of a process project moves to the recognition of whole word by combining two components, for which we achieved high accuracy, based on the context of a sentence and a lexicon.

# REFERENCES

[1] K. Riaz, "Baseline for urdu ir evaluation," in *Proceedings of the 2nd ACM workshop on Improving non english web searching.* ACM, 2008, pp. 97–100.

[2] S. B. Ahmed, S. Naz, S. Swati, and M. I. Razzak, "Handwritten urdu character recognition using 1-dimensional blstm classifier," *arXiv preprint arXiv:1705.05455*, 2017.

[3] F. Adeeba and S. Hussain, "Experiences in building urdu wordnet," in *Proceedings of the 9th workshop on Asian language resources*, 2011, pp. 31–35.

[4] M. Akram and S. Hussain, "Word segmentation for urdu ocr system," in *Proceedings of the 8th Workshop on Asian Language Resources, Beijing, China*, 2010, pp. 88–94.

[5] N. Durrani and S. Hussain, "Urdu word segmentation," in *Human Language Technologies: The 2010 Annual Conference of the North American Chapter of the Association for Computational Linguistics.* Association for Computational Linguistics, 2010, pp. 528–536.

[6] K. Wu, E. Otoo, and K. Suzuki, "Optimizing two-pass connected-component labeling algorithms," vol. 12, no. 2. Springer, 2009, pp. 117–135.

[7] K. Khan, R. Ullah, N. A. Khan, and K. Naveed, "Urdu character recognition using principal component analysis," *International Journal of Computer Applications*, vol. 60, no. 11, 2012.

[8] N. Sabbour and F. Shafait, "A segmentation-free approach to arabic and urdu ocr." in *DRR*, 2013, p. 86580N.

[9] I. Ahmad, X. Wang, R. Li, and S. Rasheed, "Offline urdu nastaleeq optical character recognition based on stacked denoising autoencoder," *China Communications*, vol. 14, no. 1, pp. 146–157, 2017.

[10] S. Naz, A. I. Umar, R. Ahmad, S. B. Ahmed, S. H. Shirazi, and M. I. Razzak, "Urdu nastaliq text recognition system based on multi-dimensional recurrent neural network and statistical features," *Neural Computing and Applications*, vol. 28, no. 2, pp. 219–231, 2017.

[11] S. Naz, A. I. Umar, R. Ahmad, I. Siddiqi, S. B. Ahmed, M. I. Razzak, and F. Shafait, "Urdu nastaliq recognition using convolutional–recursive deep learning," *Neurocomputing*, vol. 243, pp. 80–87, 2017.

[12] N. Shahzad, B. Paulson, and T. Hammond, "Urdu qaeda: recognition system for isolated urdu characters," in *Proceedings of the IUI Workshop on Sketch Recognition, Sanibel Island, Florida*, 2009.

[13] M. W. Sagheer, C. L. He, N. Nobile, and C. Y. Suen, "Holistic urdu handwritten word recognition using support vector machine," in *Proceedings of the 2010 20th International Conference on Pattern Recognition.* IEEE Computer Society, 2010, pp. 1900–1903.

[14] I. Ahmad, X. Wang, Y. hao Mao, G. Liu, H. Ahmad, and R. Ullah, "Ligature based urdu nastaleeq sentence recognition using gated bidirectional long short term memory," *Cluster Computing*, pp. 1–12, 2017.

[15] N. H. Khan, A. Adnan, and S. Basar, "Urdu ligature recognition using multi-level agglomerative hierarchical clustering," *Cluster Computing*, pp. 1–12, 2017.

[16] M. Blumenstein and B. Verma, "Neural-based solutions for the segmentation and recognition of difficult handwritten words from a benchmark database," in *Document Analysis and Recognition, 1999. ICDAR'99. Proceedings of the Fifth International Conference on.* IEEE, 1999, pp. 281–284.

[17] Z. Ahmad, J. K. Orakzai, and I. Shamsher, "Urdu compound character recognition using feed forward neural networks," in *Computer Science and Information Technology, 2009. ICCSIT 2009. 2nd IEEE International Conference on.* IEEE, 2009, pp. 457–462.

[18] T. Ali, T. Ahmad, and M. Imran, "Uocr: A ligature based approach for an urdu ocr system," in *Computing for Sustainable Global Development (INDIACom), 2016 3rd International Conference on.* IEEE, 2016, pp. 388–394.

[19] M. Kavianifar and A. Amin, "Preprocessing and structural feature extraction for a multi-fonts arabic/persian ocr," in *Document Analysis and Recognition, 1999. ICDAR'99. Proceedings of the Fifth International Conference on.* IEEE, 1999, pp. 213–216.

[20] S. S. Ahranjany, F. Razzazi, and M. H. Ghassemian, "A very high accuracy handwritten character recognition system for farsi/arabic digits using convolutional neural networks," in *Bio-Inspired Computing: Theories and Applications (BIC-TA), 2010 IEEE Fifth International Conference on.* IEEE, 2010, pp. 1585–1592.

[21] M. Zhang, Y. Zhang, and G. Fu, "Transition-based neural word segmentation." 2016.

[22] Y. LeCun, Y. Bengio, and G. Hinton, "Deep learning," *Nature*, vol. 521, no. 7553, pp. 436–444, 2015.

[23] K. He, X. Zhang, S. Ren, and J. Sun, "Deep residual learning for image recognition," in *Proceedings of the IEEE conference on computer vision and pattern recognition*, 2016, pp. 770–778.

[24] Q. Akram, S. Hussain, F. Adeeba, S. Rehman, and M. Saeed, "Framework of urdu nastalique optical character recognition system," in *the Proceedings of Conference on Language and Technology.(CLT 14), Karachi, Pakistan*, 2014.

[25] B. Epshtein, E. Ofek, and Y. Wexler, "Detecting text in natural scenes with stroke width transform," in *Computer Vision and Pattern Recognition (CVPR), 2010 IEEE Conference on.* IEEE, 2010, pp. 2963–2970.

[26] D. A. Satti and K. Saleem, "Complexities and implementation challenges in offline urdu nastaliq ocr," in *Proceedings of the Conference on Language and Technology*, 2012, pp. 85–91.